# AutoDRIVE Simulator

A Simulator for Scaled Autonomous Vehicle Research and Education

Tanmay Vilas Samak and Chinmay Vilas Samak

2020

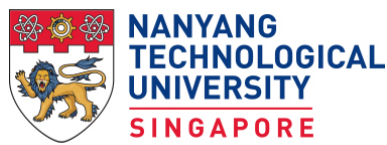

# ABSTRACT

AutoDRIVE is envisioned to be a comprehensive research platform for scaled autonomous vehicles. This work is a stepping-stone towards the greater goal of realizing such a research platform. Particularly, this work proposes a pseudo-realistic simulator for scaled autonomous vehicles, which is targeted towards simplicity, modularity and flexibility. The AutoDRIVE Simulator not only mimics realistic system dynamics but also simulates a comprehensive sensor suite and realistic actuator response. The simulator also features a communication bridge in order to interface externally developed autonomous driving software stack, which allows users to design and develop their algorithms conveniently and have them tested on our simulator. Presently, the bridge is compatible with Robot Operating System (ROS) and can be interfaced directly with the Python and C++ scripts developed as a part of this project. The bridge supports local as well as distributed computing.

# ACKNOWLEDGEMENT

At the commencement of this report, we would like to add a few words of appreciation for all those who have been a part of this project; directly or indirectly.

We would like to express our immense gratitude towards Nanyang Technological University and the NTU-India Connect Office for providing us this research internship opportunity, as a whole, and for providing us with an excellent atmosphere for working on this project.

Our sincere thanks also go to the Department of Mechanical and Aerospace Engineering, Nanyang Technological University for providing all facilities and support to meet our project requirements.

We would also like to express our deep and sincere gratitude to our project guide Dr. Xie Ming for his valuable guidance, consistent encouragement, personal caring and timely help. All through the work, in spite of his busy schedule, he has extended cheerful and cordial support to us for completing this work and making this project a success. His guidance helped us in all the time of project and writing of this report. We could not have imagined having a better advisor and mentor for this project.

Finally, we are grateful to all the sources of information without which this project would be incomplete. It is due to their efforts and research that our report is more accurate and convincing.

<div align="right">Authors</div>

# TABLE OF CONTENTS

# LIST OF TABLES

# LIST OF FIGURES

# 1. INTRODUCTION

Autonomous vehicles have been a dream of mankind for many years and the field has witnessed an accelerated advancement over the past few decades owing to the continuous and focused research and development. However, this demands for rapid development and testing of various aspects of vehicle autonomy through novel algorithm design, which in turn demands for design and development of efficient tools aiding the researches throughout the process.

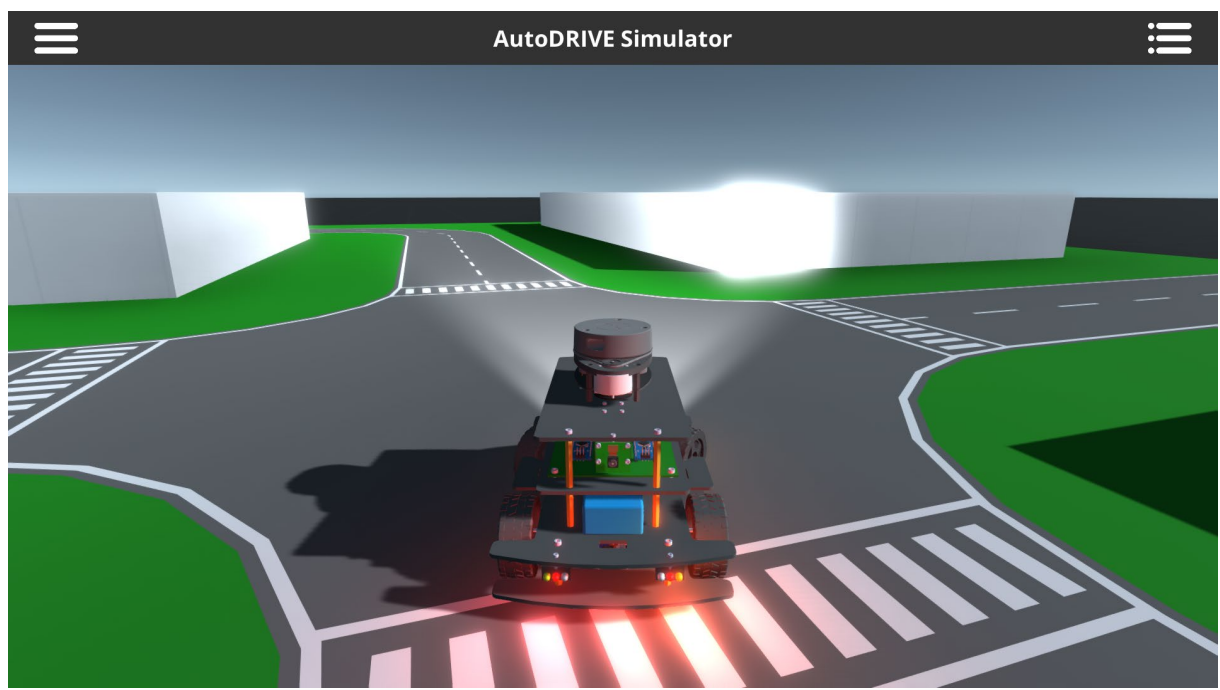

Figure 1. The AutoDRIVE Simulator

In this technical report, we present AutoDRIVE Simulator [1], a cross-platform simulator for scaled autonomous vehicle research. The simulator is developed atop the Unity game engine, which enables simulation of realistic physics (using NVIDIA's PhysX engine) as well as photorealistic graphics (using Unity's Post-Processing Stack). The simulator can be exploited by the users (particularly targeting students and researchers in the field) in order to develop and test their algorithms aimed at autonomous driving.

## 1.1 Motivation

The present autonomous driving research community lacks tools for design and development of scaled autonomous vehicles. Scaling down the task of autonomous driving implies reduced cost and accelerated prototyping. Such a scaled research platform shall open up an avenue for academic education and research.

## 1.2 Objectives

This research project was aimed at developing a simulator for robotics and autonomous systems. In particular, we chose to address the task of development of a simulation system for scaled autonomous vehicles.

## 1.3 Scope

The scope of the presented simulation system includes design and development of a virtual vehicle (including a comprehensive sensor suite and actuators), a modular environment, a communication bridge and a convenient user interface. It is to be noted that the development of the autonomous driving software stack is left to the users and that sufficient interfacing tools are provided for this purpose.

## 2. LITERATURE SURVEY

Simulators are a key value addition to the research phase as they allow virtual prototyping of the system and surrounding under study. They ensure physical safety and enable rapid prototyping at minimal costs. They also allow simulation of critical situations and corner-cases which are rarely encountered in real-world along with permutations and combinations of various sub-scenarios. Furthermore, reproduction of scenario-specific implementations is also possible in a simulated environment.

The automotive industry has long practiced the use of simulators to simulate near-realistic vehicle dynamics, specifically for rapid design and development of automotive sub-systems. ANSYS [2] and ADAMS [3] are prime examples of commercial simulators adopted by industry and academia for simulating realistic dynamics for vehicle design and testing. However, with the advent of autonomous vehicles, advanced driver assistant systems (ADAS) in particular, most of these simulators have started releasing updates that include autonomous driving features of some form or the other. Simulators such as CarMaker [4] and CarSim [5] are two amongst a wide range of such automotive simulators that have released autonomous driving updates in the recent past.

TORCS [6] is one of the earliest open-source simulators, which focuses predominantly on the task of manual as well as autonomous racing. Gazebo [7] is one of the most popular open source simulators for robotics research and development and has been natively adopted by ROS. AirSim [8], CARLA [9], LGSVL Simulator [10] and Deepdrive [11] are some of the other open source simulators that particularly address the problem of autonomous driving.

Commercial automotive simulators focusing on autonomous driving and driver assistance systems include NVIDIA Drive Constellation [12], rFpro [13], dSPACE [14], PreScan [15], Cognata [16] and Metamoto [17]. Researchers have also adopted commercial video games such as Grand Theft Auto V [18-20] to simulate autonomous vehicles.

However, none of these simulators support scaled autonomous vehicles and environments, which is the key problem statement for this project.

## 3. METHODOLOGY

The task of simulator development was split into four distinct stages, viz. design and development of the vehicle, the environment, the communication bridge and the user interface.

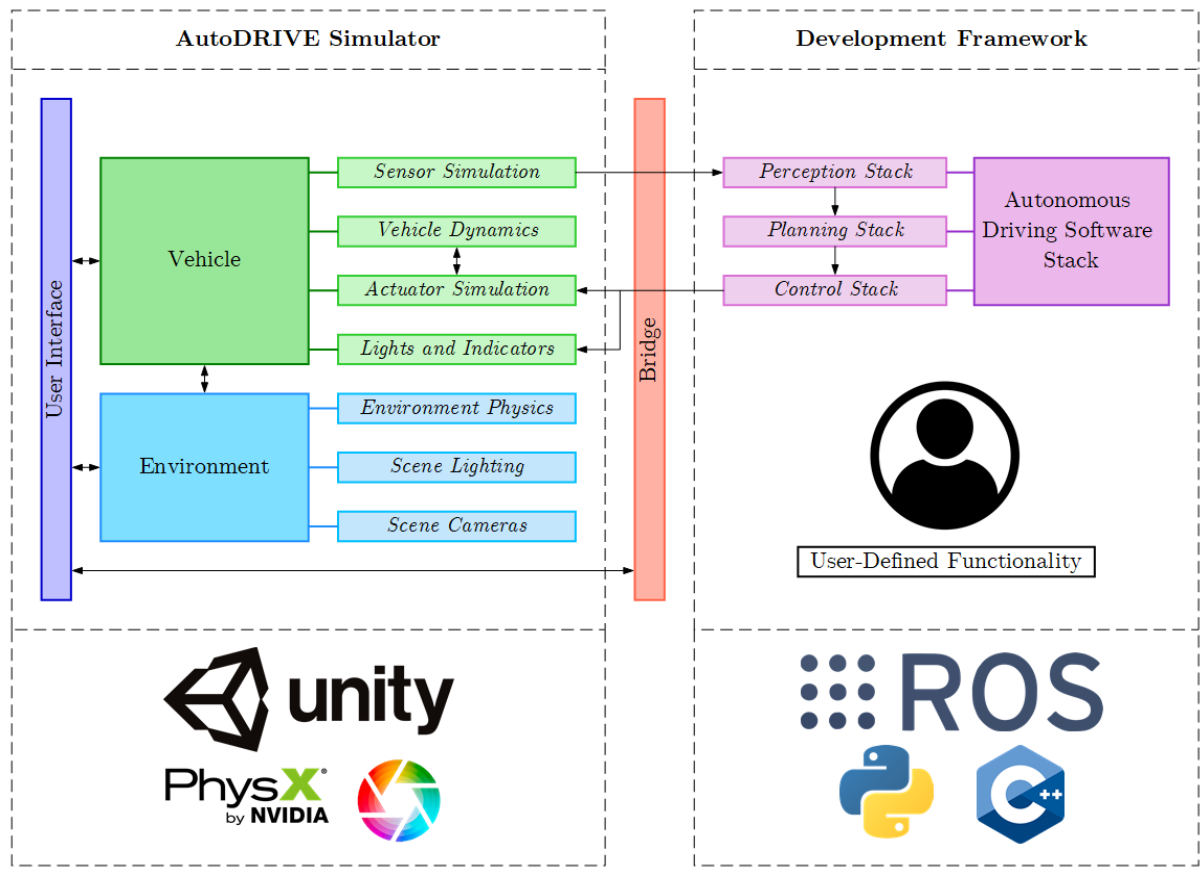

Figure 2. Simulation System Architecture

The AutoDRIVE Simulator is developed using the Unity game engine and exploits its built-in multi-threaded physics simulation engine called PhysX for simulating system dynamics. It also utilizes the Post-Processing Stack for generating photorealistic graphics.

The development framework comprises of the AutoDRIVE ROS Package as well as direct scripting support for Python and C++. As discussed earlier, the task of implementing autonomous driving software stack is outside the scope of this research project and is left to the users.

### 3.1 Vehicle

The vehicle was designed with an aim of enabling hardware development in the future, thereby achieving the second milestone of the AutoDRIVE project.

The designed vehicle is approximately a 1:14 scale rear-wheel drive model. It primarily comprises of four different platforms fastened together at appropriate offsets using standoffs. The first platform houses power electronics components and drive actuators. The second platform houses the steering actuator coupled to a steering mechanism and also serves as a place to strap on a LiPo battery. It also has provision for mounting front and rear vehicle lights.

The third platform houses an on-board computer and a custom circuit board with microcontrollers, sensors, power distribution circuit and interfacing components. Finally, the fourth platform houses the LIDAR and has provision for mounting LIDAR USB Adapter Board as well as front and rear vehicle cameras.

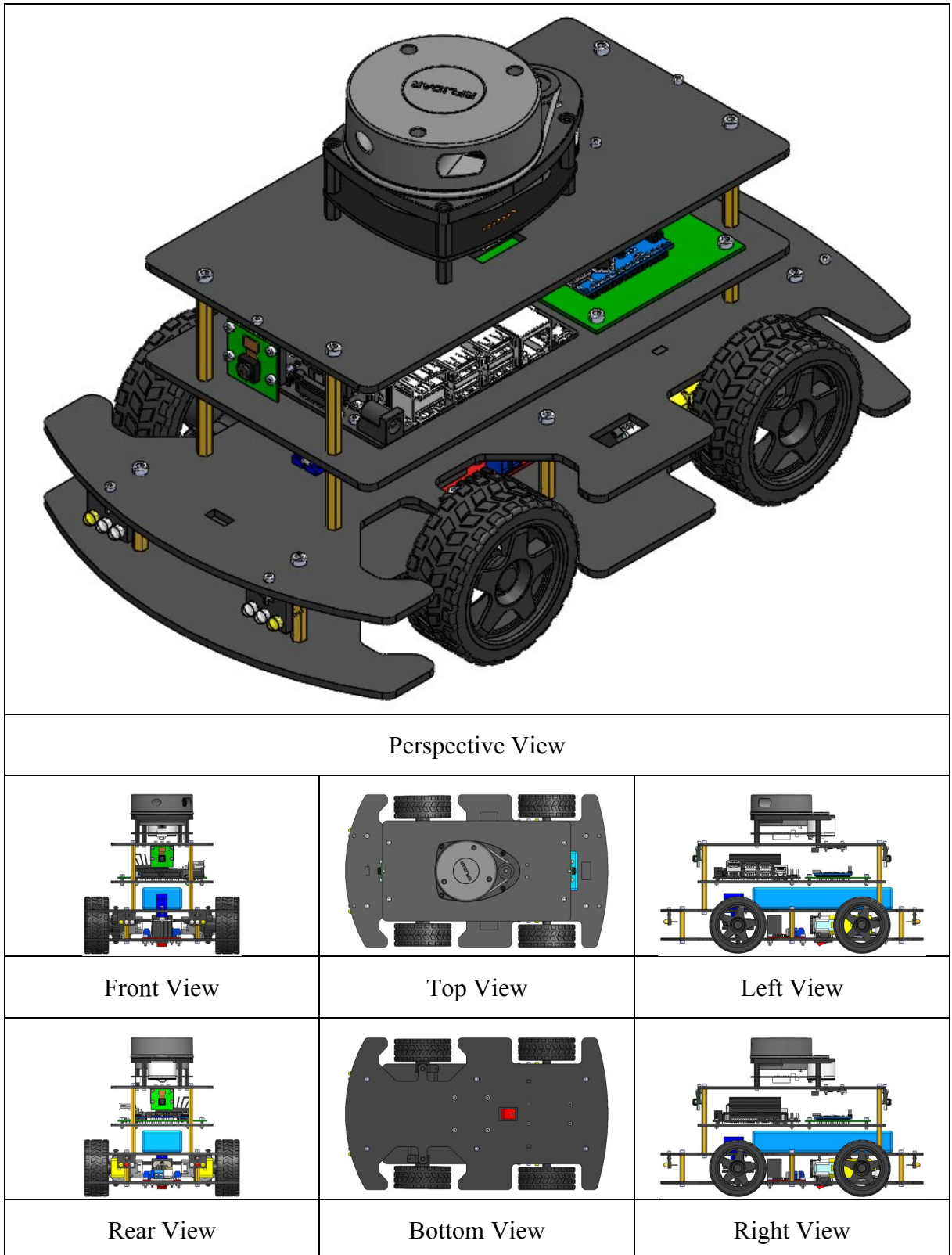

*Figure 3. Vehicle Design*

Following is a brief summary of the designed vehicle.

Table 1. Vehicle Design Summary

| OVERVIEW | |
|---|---|
| Dimensions (LxBxH) | 300x175x182 mm |
| Mass | 1.75 kg |
| Chassis | 3 mm Thick Acrylic Cutouts |
| Fixtures | Nut-Bolts, Standoffs, Couplings, etc. |
| Drive Architecture | Rear Wheel Drive (2WD), Front Wheel Steered |
| Drive Technology | Fully Electric |
| Gear | Drive, Reverse (H-Bridge) |
| Throttle | PWM Controlled Proportional Throttle |
| Brake | Automatic |
| Steering | 4-Bar Mechanism |
| Front Lights | Head Lights, Turning Indicators |
| Rear Lights | Tail Lights, Turning Indicators, Reverse Indicators |
| **POWER** | |
| Li-Po Battery | 11.1 V 5200 mAh 25 C |
| Buck Converter | 5 V 5 A DC-DC Buck Converter |
| Motor Driver | L298N |
| **SENSORS** | |
| Throttle Sensor | Virtual Sensor |
| Steering Angle Sensor | Virtual Sensor |
| Motor Encoders | 16 PPR Magnetic Rotary Encoder |
| IPS | Virtual Sensor |
| 9-Axis IMU | MPU9250 |
| LIDAR | RPLIDAR A1M8 |
| Front and Rear-View Cameras | Raspberry Pi Camera V1.3 |
| **PROCESSORS** | |
| Computer | Jetson Nano B01 |
| Microcontrollers | Arduino Nano Rev.3 |
| **ACTUATORS** | |
| Drive Actuators | 6 V TT Motors (160 RPM)* |
| Steering Actuator | SG90 |

*The drive actuators shall be operated at 5 V (130 RPM)

The vehicle was designed using SolidWorks, which is a parametric modelling software. However, Unity does not accept parametric models as assets. Hence, a conversion was to be made from the parametric model to a mesh model. This was accomplished by employing 3DS Max as an intermediate software to convert the SolidWorks Assembly file (SLDASM) into a Filmbox file (FBX).

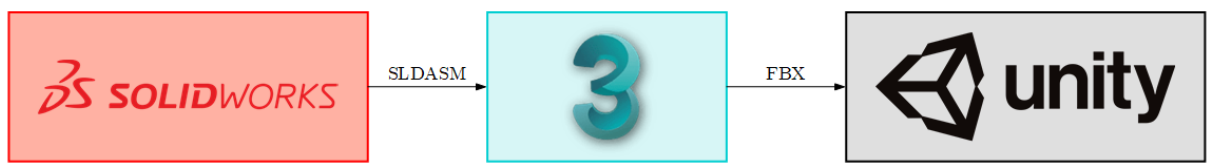

Figure 4. Vehicle Asset Development Workflow

### 3.1.1 Sensor Simulation

The simulator currently supports seven different sensors, viz. throttle sensor, steering sensor, motor encoders, Indoor Positioning System (IPS), Inertial Measurement Unit (IMU), LIDAR and cameras.

**Throttle Sensor**

The throttle sensor is a virtual sensor, which measures the instantaneous throttle value of the vehicle. In reality, it records the latest throttle command sent to the drive actuators. While positive readings of the throttle sensor indicate forward driving, negative readings indicate reverse driving. A zero-throttle reading implies that the vehicle is in a stand-still condition as the brakes are automatically applied. Following are specifications of the throttle sensor.

*Table 2. Throttle Sensor Specifications*

| THROTTLE SENSOR | |
|---|---|
| Type | Virtual Sensor |
| Class | Actuator Feedback |
| Supported Outputs | [-1, 1] |

**Steering Sensor**

The steering sensor is a virtual sensor, which measures the instantaneous steering angle of the vehicle. In reality, it records the latest steering command sent to the steering actuator. While positive readings of the steering sensor indicate right turns, negative readings indicate left turns. A zero-steering reading implies that the vehicle is driving straight. Following are specifications of the steering sensor.

*Table 3. Steering Sensor Specifications*

| STEERING SENSOR | |
|---|---|
| Type | Virtual Sensor |
| Class | Actuator Feedback |
| Supported Outputs | [-1, 1] |

**Motor Encoders**

The drive actuators are each installed with a simulated incremental encoder. The simulated encoders measure the angular displacement of the motor shaft (and hence the wheels) and update the count once the wheel has turned by a predefined amount. They not only output the pulse count but also directly provide angle turned by the wheel. Following are specifications of the motor encoders.

*Table 4. Motor Encoder Specifications*

| MOTOR ENCODER | |
|---|---|
| Type | Simulated Sensor |
| Class | Proprioceptive |
| Pulses Per Revolution (PPR) | 16 |
| Supported Outputs | Ticks, Angle |

## IPS

The IPS is a scaled form of Global Navigation Satellite System (GNSS). It locates the vehicle within the map. Following are specifications of the IPS.

*Table 5. IPS Specifications*

| IPS | |
|---|---|
| Type | Simulated Sensor |
| Class | Proprioceptive |
| Supported Outputs | Position Vector [x, y, z] m |

## IMU

The IMU measures inertial data of the vehicle. This includes the vehicle's orientation (measured using a magnetometer), angular velocity (measured using a gyroscope) and linear acceleration (measured using an accelerometer). The simulated IMU converts inertial readings from Unity's native left-handed co-ordinate system into a right-handed co-ordinate system with X-axis pointing to the front of the vehicle, Y-axis pointing to the left of the vehicle and Z-axis pointing to the top of the vehicle. This conversion helps simulate realistic sensor data. Following are specifications of the IMU.

*Table 6. IMU Specifications*

| IMU | |
|---|---|
| Type | Simulated Sensor |
| Class | Proprioceptive |
| Supported Outputs | Orientation Quaternion [x, y, z, w] |
| | Orientation Euler Angles [x, y, z] rad |
| | Angular Velocity Vector [x, y, z] rad/s |
| | Linear Acceleration Vector [x, y, z] m/s$^2$ |

## LIDAR

The Light Detection and Ranging (LIDAR) sensor measures the relative distance of objects in the scene by illuminating them with a laser beam and measuring the reflection with a sensor. The simulator implements a 2D LIDAR, which performs a 360° scan around the vehicle. Following are specifications of the LIDAR.

*Table 7. LIDAR Specifications*

| LIDAR | |
|---|---|
| Type | Simulated Sensor |
| Class | Exteroceptive |
| Scan Rate | 7 Hz |
| Measurements Per Scan | 360 |
| Minimum Range | 0.15 m |
| Maximum Range | 12 m |
| Intensity | 47 |
| Supported Outputs | Range Array, Intensity Array |

#### Cameras

Cameras act as *"eyes"* for the vehicle and provide visual sensory data in form of image frames. The simulated vehicle has front and rear-view cameras, which assist in forward as well as reverse driving. Following are specifications of the cameras.

*Table 8. Camera Specifications*

| CAMERA | |
|---|---|
| Field of View | 41.669° |
| Target Resolution | 640x480p |
| Focal Length | 3.6 mm |
| Sensor Size | 3.76x2.74 mm |

### 3.1.2 Vehicle Dynamics

The notion of physical modelling was adopted in order to develop the dynamic model of the vehicle and Unity's inbuilt PhysX engine was used for real time computation and updation of the vehicle motion based on the control inputs and its resulting interaction with the environment.

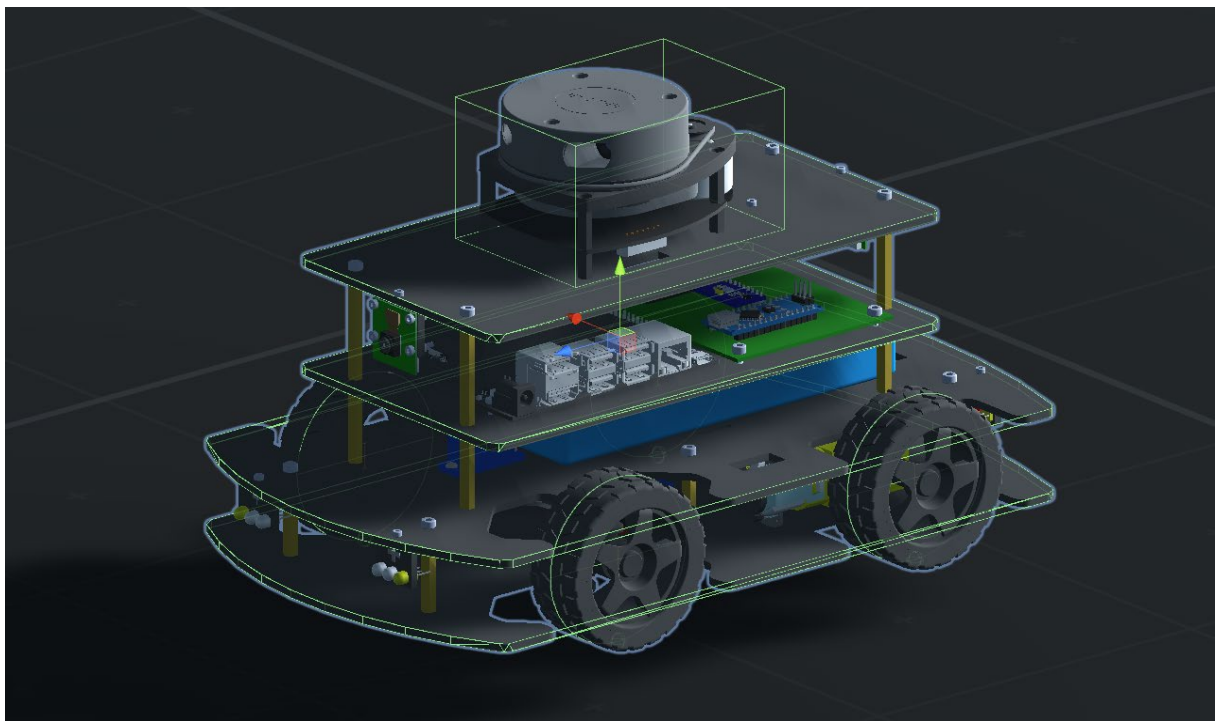

*Figure 5. Physical Model of the Vehicle in Unity Scene*

The developed vehicle model has various physical components associated with it, such as a rigid body to simulate gravity and other forces acting on the vehicle, body colliders to simulate realistic collisions and, most importantly, wheel colliders to simulate realistic driving behaviour. In addition to this, all the system parameters are configured to simulate realistic vehicular dynamics. The following table summarizes system parameters for the simulated vehicle.

*Table 9. System Parameters of Vehicle Model*

| RRIGID BODY | | |
|---|---|---|
| Mass | | 1.75 kg |
| Drag | | 0.1 N |
| Angular Drag | | 0.05 N |
| **MESH COLLIDERS** | | |
| Mesh Collider 1 | | Platform 1 |
| Mesh Collider 2 | | Platform 2 |
| Mesh Collider 3 | | Platform 3 |
| Mesh Collider 4 | | Platform 4 |
| **BOX COLLIDERS** | | |
| Box Collider 1 | | LIDAR |
| **WHEEL COLLIDERS** | | |
| Mass | | 0.034 kg |
| Radius | | 0.0325 m |
| Damping Rate | | 0.025 |
| Suspension Distance | | 0.01 m |
| Force Application Point Distance | | 0 m |
| Centre | | [0, 0.005, 0] m |
| Suspension | Spring Coefficient | 5040 N/m |
| | Damper Coefficient | 20 kg/s |
| | Target Position | 0.5 |
| Longitudinal Friction | Extremum Slip | 0.4 |
| | Extremum Value | 1 |
| | Asymptote Slip | 0.8 |
| | Asymptote Value | 0.5 |
| | Stiffness | 1 |
| Lateral Friction | Extremum Slip | 0.2 |
| | Extremum Value | 1 |
| | Asymptote Slip | 0.5 |
| | Asymptote Value | 0.75 |
| | Stiffness | 1 |

### 3.1.3 Actuator Simulation

The vehicle actuation system consists of drive actuators (for longitudinal motion) and steering actuator (for lateral actuation). While the drive actuators drive the rear wheels at a specified speed, the steering actuator steers the front wheels at a specified angle.

In order to simulate the high holding torque of the drive actuators, brakes are automatically applied unless the throttle input is non-zero. Furthermore, the steering actuator is retracted to 0° unless any valid steering control input is given to the vehicle.

The actuators accept normalized control inputs so as to remove their dependency over the vehicle geometry and dynamics. All the actuators have been defined with realistic saturation limits – steering actuation limit of ±30° and drive actuation limit of 130 RPM. Furthermore, actuator response delays have also been modelled in order to match realistic actuator dynamics.

### 3.1.4 Lights and Indicators

The vehicle has fully functional headlights (disabled, low-beam, high-beam), taillights/brake indicators (disabled, partially luminous when headlights are illuminated, bright when brakes are applied), turning indicators (disabled, left indicator, right indicator, hazard indicator) and reverse indicators (disabled in drive gear, luminous in reverse gear).

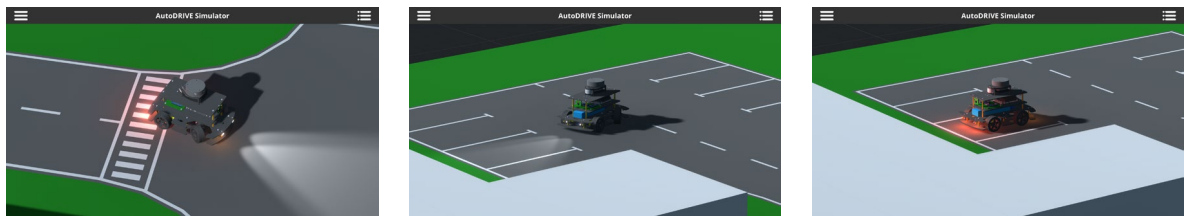

*Figure 6. Vehicle Lighting*

While headlights and turning indicators can be controlled in manual as well as autonomous mode, brake and reverse indicators are controlled automatically upon respective event detection.

### 3.2 Environment

AutoDRIVE is aimed at scaled autonomous vehicle research. As a result, the environment is also scaled down appropriately so as to match physical dimensions of the vehicle.

The central idea of environment design is to have a finite number of modules, enabling modular and reconfigurable construction of the scene, where each module is classified as drivable or non-drivable segment w.r.t. the vehicle.

The currently designed modules support dual-lane roads with white lane markings (solid at road boundaries and dashed at lane separation), right-angle turns, 3-way and 4-way intersections (with stop-lines and pedestrian crossings marked in white colour), green lawn representing non-drivable segments and a modular construction box to build 3D obstructions.

Definition of road curvature required knowledge of turning radius of the vehicle, which was calculated based on its kinematics (using kinematic bicycle model), considering a wheelbase of 141.54 mm and a total chassis length of 300 mm, with a steering actuation limit of $\pm 30°$. The following table summarizes turning radii of the vehicle w.r.t. different components.

*Table 10. Vehicle Turning Radii*

| Reference Component | Turning Radius |
|---|---|
| Front Wheel | 283.08 mm |
| Rear Wheel | 245.15 mm |
| Chassis | 600.00 mm |

The minimum road curvature was therefore defined to be 600 mm. This shall allow the entire vehicle chassis to stay within the lane while taking a turn.

AutoDRIVE Simulator currently supports three road modules, two intersection modules, two parking modules, one terrain module and one obstruction module. While most of the modules are 2D in nature, the obstruction module is one and only 3D module in the entire set.

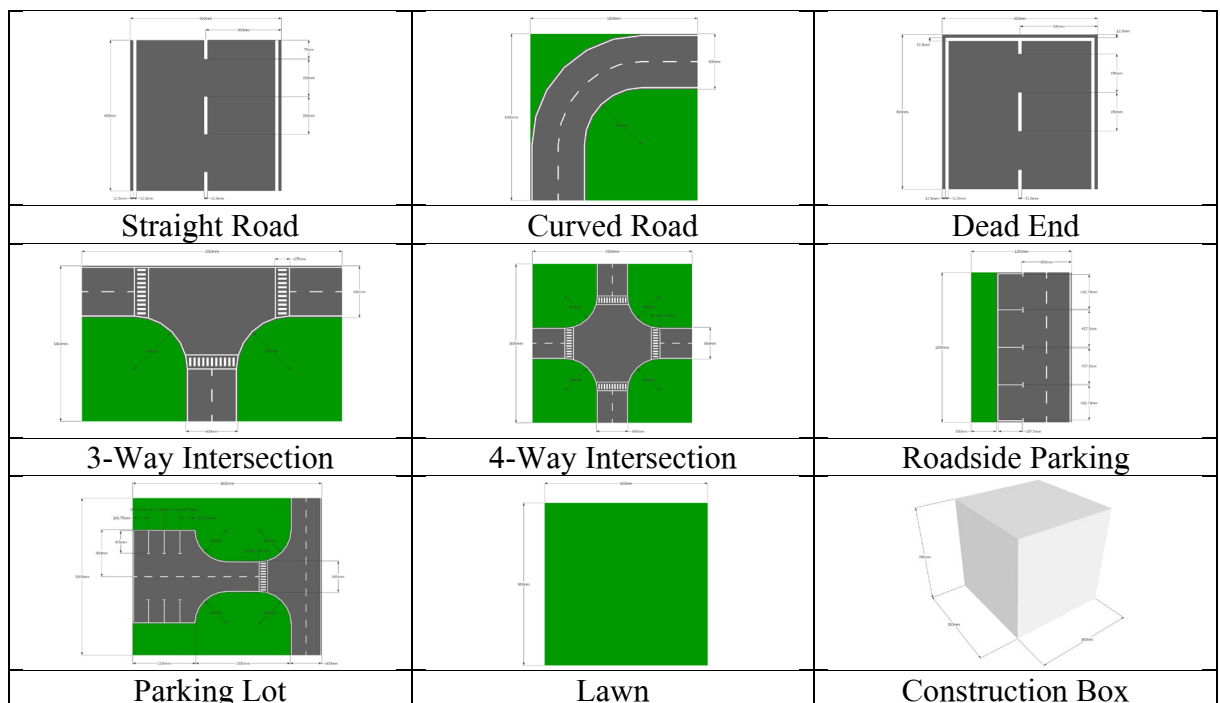

| Straight Road | Curved Road | Dead End |
| 3-Way Intersection | 4-Way Intersection | Roadside Parking |
| Parking Lot | Lawn | Construction Box |

*Figure 7. Environment Modules*

The following table summarizes each of the environment modules.

*Table 11. Environment Modules*

| Module Name | Module Type | Segment Type | Notes |
|---|---|---|---|
| Straight Road | Road | Drivable | Dual-Lane Road |
| Curved Road | Road | Drivable | Right-Angle Turn |
| Dead End | Road | Drivable | Dead End |
| 3-Way Intersection | Intersection | Drivable | T-Intersection |
| 4-Way Intersection | Intersection | Drivable | X-Intersection |
| Roadside Parking | Parking | Drivable | Parallel Parking |
| Parking Lot | Parking | Drivable | Structured Parking |
| Lawn | Terrain | Non-Drivable | Off-Road Terrain |
| Construction Box | Obstruction | Non-Drivable | 3D Obstruction |

The environment modules were developed using SketchUp, which is natively supported by Unity. This allowed direct asset creation using the SketchUp file (SKP).

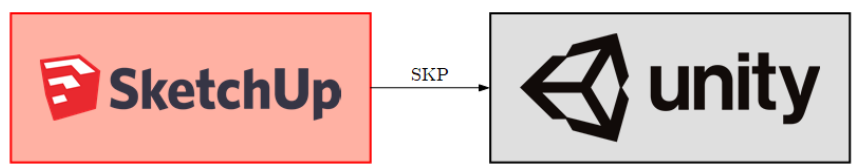

*Figure 8. Environment Asset Development Workflow*

The environment modules imported into Unity were used to create a small map. The key focus was to encompass all the modules into the map, while also minimizing space. An additional requirement was to include at least one closed loop path in the map. Following is a mini-map created by adhering to the aforementioned constraints.

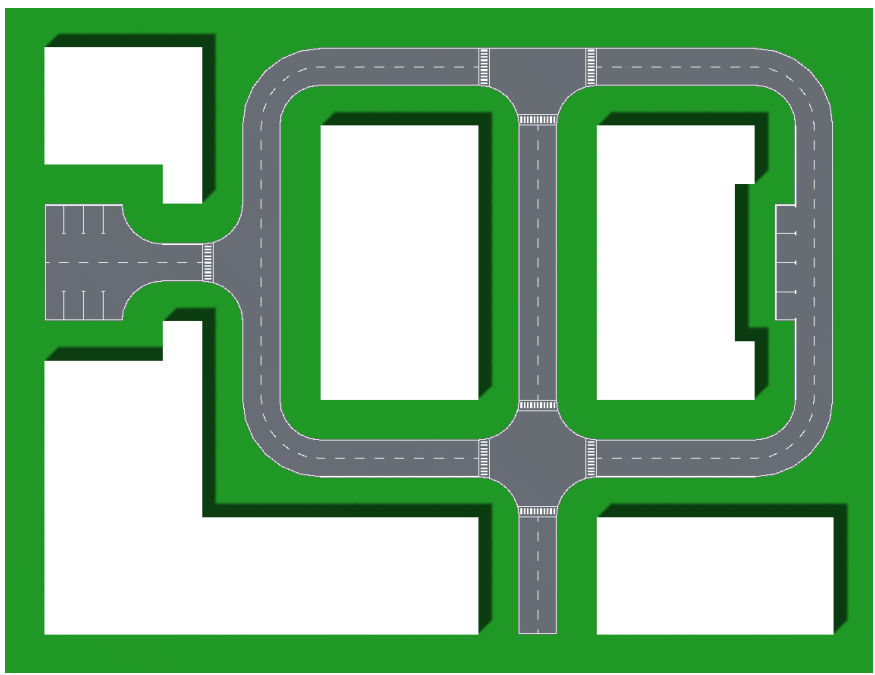

*Figure 9. Mini-Map*

### 3.2.1 Environment Physics

The 2D environment modules have a box collider attached to them and are defined to be *"static"* in nature. This implies that any interactive forces between the vehicle wheels and these modules shall only affect dynamics of the vehicle.

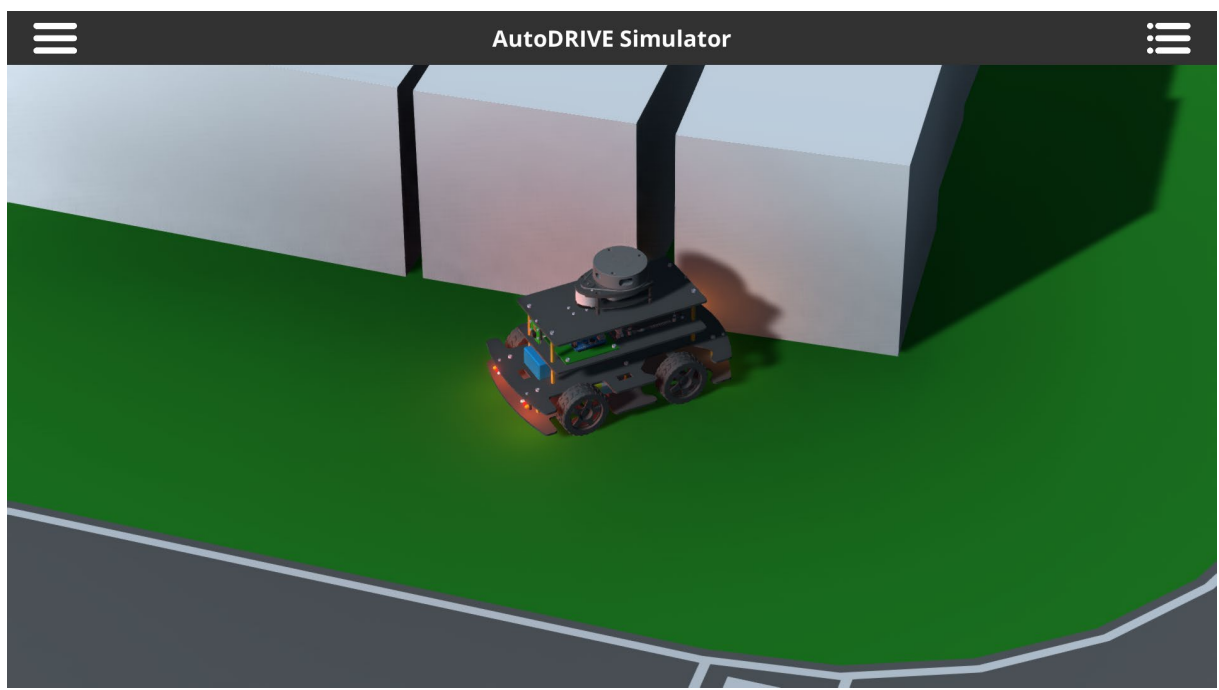

*Figure 10. Vehicle Collision Demonstration*

The 3D modules (i.e. construction boxes) also have a box collider attached to them. These modules are defined to be *"dynamic"* objects, which implies that any interaction between the vehicle and these modules shall affect dynamics of both the entities. Furthermore, interaction between these modules is also possible.

### 3.2.2 Scene Lighting

Scene lighting refers to the environmental illumination, which not only but also affects visual sensors such as cameras.

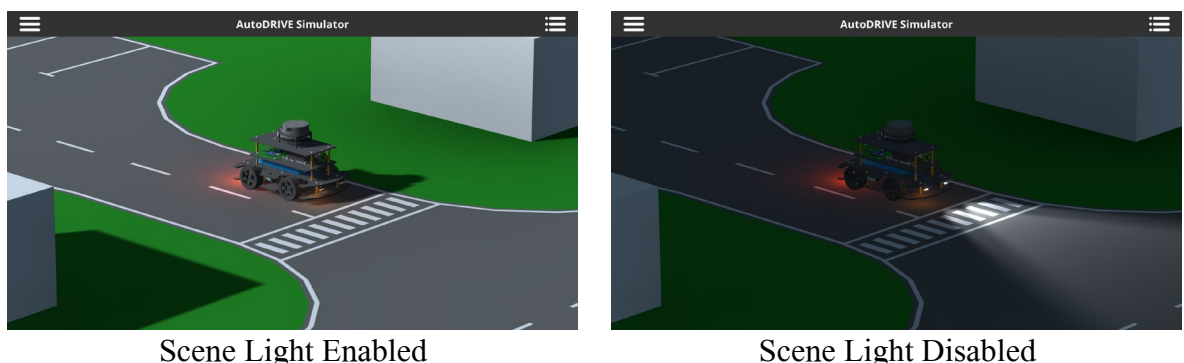

| Scene Light Enabled | Scene Light Disabled |

Figure 11. Scene Lighting

The scene light can be enabled to simulate daylight driving conditions and can be disabled to simulate night driving conditions.

### 3.2.3 Scene Cameras

Presently, the simulator supports three scene cameras, each of which provides a different view.

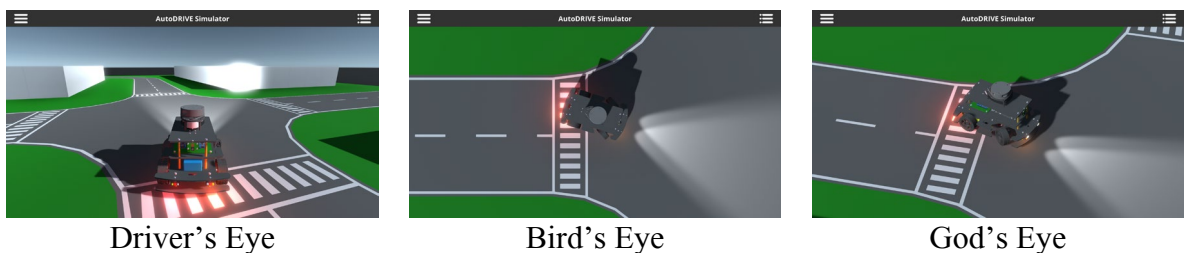

Driver's Eye    Bird's Eye    God's Eye

Figure 12. Scene Cameras

**Driver's Eye:** This camera continuously follows the vehicle from the rear side and its transform is static w.r.t. the vehicle. This view is convenient for teleoperating the vehicle manually or to follow it, in general.

**Bird's Eye:** This camera continuously tracks the vehicle from a top view (without rotating) and has a provision to zoom in or out as required in order to give the user a better visualization of the vehicle within the environment. This view is convenient for driving the vehicle manually (especially in reverse direction) or for scenic visualization/recording in autonomous driving mode.

**God's Eye:** This camera continuously tracks the vehicle while automatically adjusting its focus and field of view for efficient visualization. This view is especially useful for scenic visualization/recording in autonomous driving mode.

### 3.3 Communication Bridge

The communication bridge is implemented using WebSocket, which provides full-duplex communication channels over a single TCP connection. Another peculiar advantage of WebSocket is that it allows event-driven responses, which significantly enhances the bridge functionality. Finally, WebSocket also minimizes overhead per message, thereby making it an efficient protocol for our application.

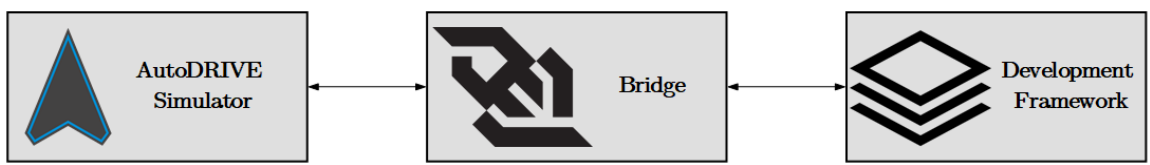

*Figure 13. Communication Bridge Layout*

The bridge parameters are reconfigurable, thereby allowing users to enter specific IP address and port number in order to establish a server-client communication between their scripts (server) and the simulator (client). This allows local as well as distributed networking, meaning users can run the simulator and scripts on the same machine (configured as default bridge parameters 127.0.0.1:4567) or on different machines connected through a common network (LAN/WLAN) by specifying custom bridge parameters. This expands the computation limits of the simulation system.

## 3.4 User Interface

AutoDRIVE Simulator aims at providing supreme user experience. There are two fundamental components of the user interface viz. hardware input methods and graphical user interface (GUI).

### 3.4.1 Hardware User Interface

The simulator currently accepts input from hardware interfaces including keyboard and mouse. In addition to interacting with the GUI, the hardware inputs also serve as key control elements for the simulated vehicle. Following table summarizes the hardware inputs and their respective functionalities.

*Table 12. Hardware User Interface*

| Hardware | User Input | Function |
| --- | --- | --- |
| Keyboard | W/Up Arrow | Drive Vehicle Forward |
| Keyboard | S/Down Arrow | Drive Vehicle Reverse |
| Keyboard | A/Left Arrow | Steer Vehicle Left |
| Keyboard | D/Right Arrow | Steer Vehicle Right |
| Keyboard | G | Toggle Headlights (Low Beam) |
| Keyboard | H | Toggle Headlights (High Beam) |
| Keyboard | L | Toggle Left Turn Indicator |
| Keyboard | R | Toggle Right Turn Indicator |
| Keyboard | E | Toggle Hazard Indicator |
| Keyboard | A-Z, 0-9 and Special Characters | Input Data through GUI |
| Mouse | Click | Interact with GUI |
| Mouse | Scroll | Zoom (Bird's Eye View) |

### 3.4.2 Graphical User Interface

The graphical user interface (GUI) consists of a toolbar encompassing two main panels, namely Menu and Heads-Up Display (HUD). Both the panels can be enabled/disabled using the buttons provided on the toolbar. Following is the summary of elements included within these two panels.

*Table 13. Graphical User Interface*

| **MENU PANEL** | |
|---|---|
| IP Address Field | Input field to specify IP address for the bridge (default is 127.0.0.1, i.e. standard address for IPv4 loopback traffic). |
| Port Number Field | Input field to specify port number for the bridge (default is 4567). |
| Connection Button | Button to establish connection with the server (the button is disabled once the connection is established). The status of bridge connection (i.e. Connected/Disconnected) is displayed besides this button. |
| Reset Button | Button to reset the simulator to initial conditions. The bridge parameters are also reset to default values. |
| Driving Mode Button | Button to toggle the driving mode of the ego-vehicle between Manual and Autonomous (default is Manual). The driving mode is displayed besides this button. |
| Camera Button | Button to toggle the camera view between Driver's Eye, Bird's Eye and God's Eye (default is Driver's Eye). |
| Scene Light Button | Button to enable/disable the scene light (default is enabled). |
| **HUD PANEL** | |
| Simulation Parameters | *Simulation Time (HH:MM:SS)* denoting time since the start of the simulation. Reset button resets the simulation time. |
| | *Frame Rate (Hz)* denoting the running average of the FPS value. |
| Ego-Vehicle Status | *Driving Mode* denoting the driving mode of the ego-vehicle [either Manual or Autonomous]. |
| | *Gear* denoting the drive direction of ego-vehicle, either Drive (D) or Reverse (R). |
| | *Speed (m/s)* denoting the scalar speed of the ego-vehicle. It denotes magnitude of the forward velocity of the ego-vehicle. |
| Actuator Feedbacks | *Throttle (%)* denoting the throttle input of the ego-vehicle. |
| | *Steering (rad)* denoting the steering angle of the ego-vehicle. |
| Encoder Ticks | *Encoder Ticks* of the left and right rear wheels of the ego-vehicle represented using a 1D array of 2 elements [left_encoder_ticks, right_encoder_ticks]. |
| IPS Data | *Position (m)* of the ego-vehicle within the map represented using a vector [x, y, z]. The spawn location of the vehicle is set to [0, 0] in X-Y plane. |
| IMU Data | *Orientation (rad)* denoting a vector [x, y, z] containing the Euler Angle representation of the orientation of the ego-vehicle about its local axes. |
| | *Angular Velocity (rad/s)* denoting a vector [x, y, z] representing the angular velocity of the ego-vehicle about its local axes. |
| | Linear Acceleration (m/s$^2$) denoting a vector [x, y, z] representing the linear accelerations of the ego-vehicle along its local axes. |
| LIDAR Measurement | *Ranging Measurement (m)* of LIDAR at the present time-instant. |
| Camera Previews | *Front Camera Preview* denoting the raw image from the Front Camera of the ego-vehicle. |
| | *Rear Camera Preview* denoting the raw image from the Rear Camera of the ego-vehicle. |

The following screenshot illustrates both GUI panels being enabled – the left one is Menu Panel, while the right one is HUD Panel.

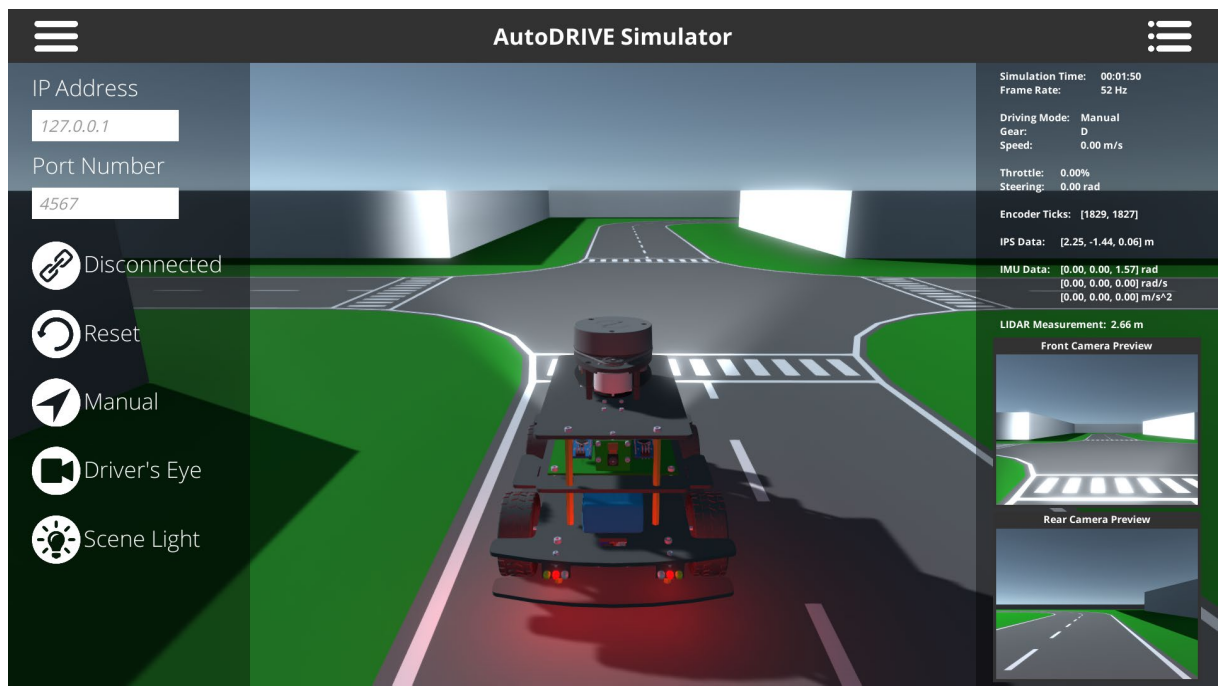

*Figure 14. Graphical User Interface*

# 4. RESULTS AND DISCUSSION

The AutoDRIVE Simulator project comprises of three key elements, namely the simulator application, a ROS package and direct scripting interfaces (Python and C++).

The standalone simulator application is targeted at Full HD (1920x1080p) resolution and supports Windows, Linux as well as Mac OS X (thanks to Unity's support for cross-platform development). It is very light-weight (~110 MB) and utilizes system resources wisely. This enables deployment of the simulator application and development framework on a single machine; nonetheless, distributed computing is also supported.

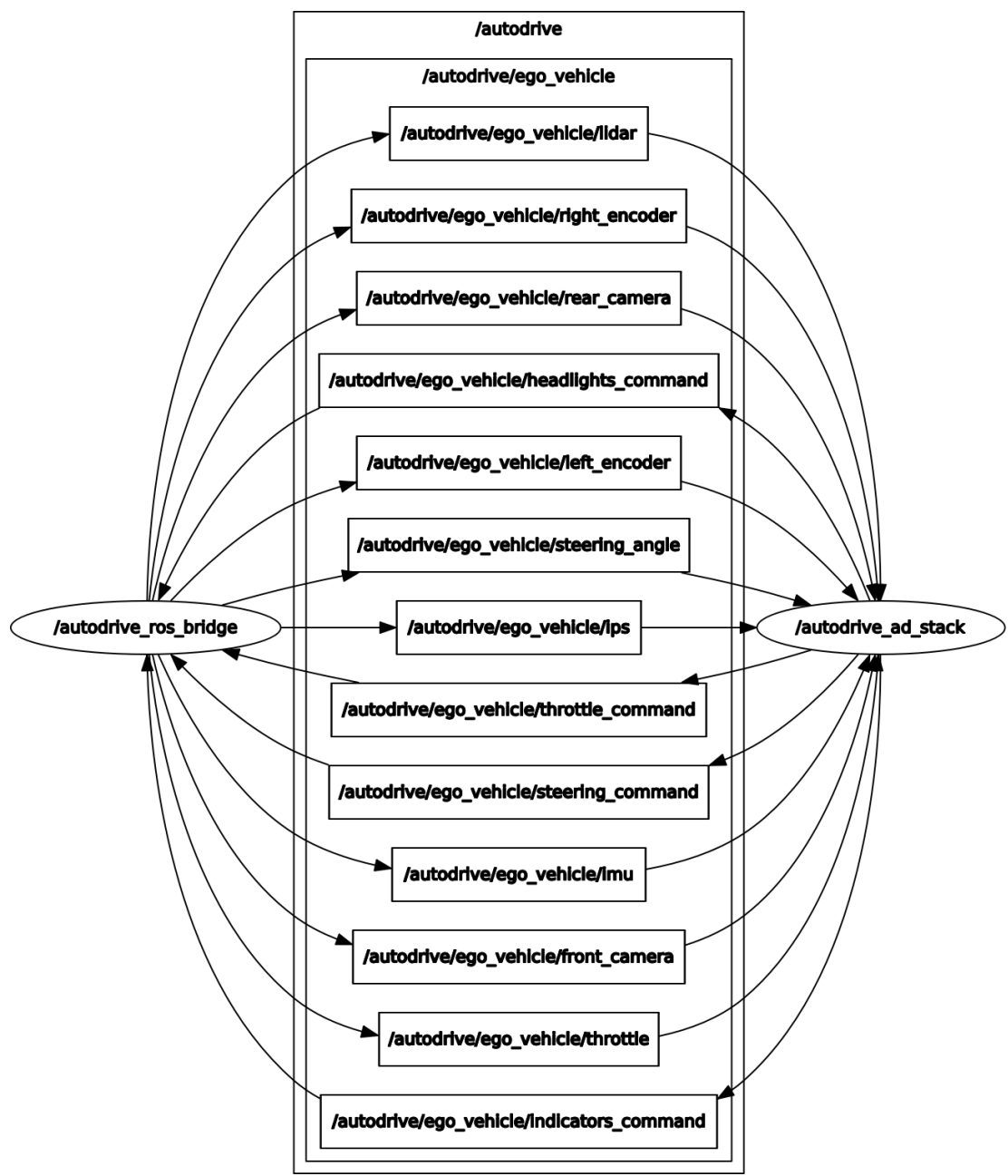

*Figure 15. ROS Computation Graph*

**Note:** *Please note that the /autodrive_ad_stack node is developed only for representational purpose and that users can implement their autonomous driving software stack as they please.*

The ROS package developed as a part of this project aims at modular development of autonomous driving software stack. The package comprises of a pre-developed ROS node called **/autodrive_ros_bridge** that handles bilateral communication with the simulator. The node receives sensor data from the simulator and publishes them as standard sensor messages on appropriate topics. The node also subscribes to the final control commands generated by the autonomous driving software stack for driving the ego-vehicle (in autonomous driving mode) and transmits those control commands to the simulator. As an icing on the cake, the ROS package also includes a pre-configured RViz file in order to visualize the ego-vehicle state and supported sensor readings on the server-side.

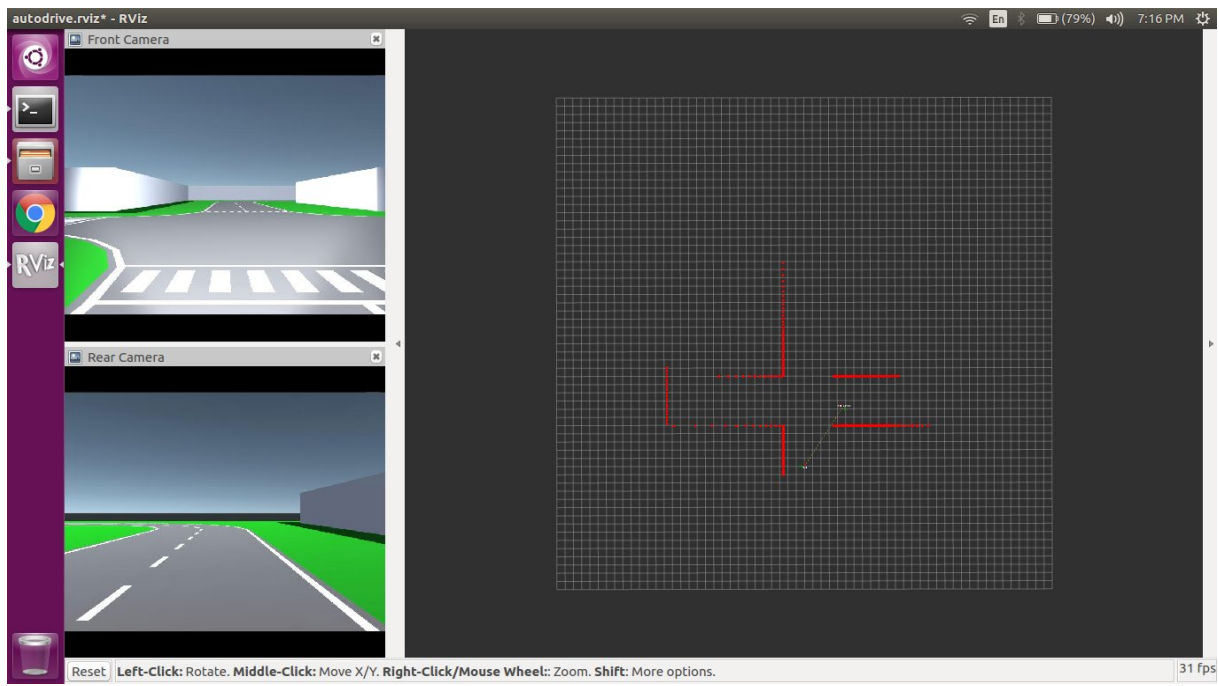

*Figure 16. ROS Visualization*

AutoDRIVE Simulator also supports general purpose scripting interfaces (Python and C++) that were developed as a part of this project. These interfaces can be used to directly communicate with the simulator without ROS (or any other middleware) as an intermediary. The scripting interfaces currently support receiving sensor data from the simulator and transmitting control commands back to it. These interfaces shall prove to be extremely efficient during the prototyping phase of various algorithms as they allow direct deployment.

# 5. CONCLUSION

In this report, we introduced AutoDRIVE Simulator, a simulation ecosystem for scaled autonomous vehicles and related applications. We also disclosed some of the prominent components of this simulation system as well as some key features it has to offer to accelerate education and research in the field of autonomous systems, autonomous driving in particular.

Students and educators can utilize this simulation platform to develop and test their autonomous driving algorithms prior to hardware implementation, which will help them reduce non-recurring engineering (NRE) cost and accelerate prototyping. This project shall also enable self-funded labs, research groups and bright students from unprivileged background to work on this exponential technology without the need of purchasing any hardware components.

This simulator already has a lot to offer so as to boost the iterative develop-test cycle for the researchers. However, further enhancement shall only make it more useful for others in the field. This simulator was developed as a research project as a part of NTU India Connect Research Internship Programme, 2020 within the stipulated duration of two months and there lies a great scope for further development of this work. We shall therefore continue contributing to this project with a primary aim of addressing the needs of students, educators and researchers.

# 6. SUPPLEMENTAL MATERIAL

Code: https://github.com/Tinker-Twins/AutoDRIVE

Video: https://youtu.be/i7R79jwnqlg